\title{Modelling the semantics of text in complex document layouts using graph transformer networks.}

\author{Thomas R. Barillot*$^{,1}$, Jacob Saks$^{1}$, Polena Lilyanova$^{1}$, Edward Torgas$^{1}$,\\
 Yachen Hu$^{1}$, Yuanqing Liu$^{1}$, Varun Balupuri$^{1}$ and Paul V. Gaskell*$^{,1}$ \\ \\
 \small{*equal contribution}, \small{$^{1}$BlackRock Inc.}}

\newcommand{\abstractText}{\noindent
Representing structured text from complex documents typically calls for different machine learning techniques,
such as language models for paragraphs and convolutional neural networks (CNNs) for table extraction,
which prohibits drawing links between text spans from different content types. 
In this article we propose a model that approximates the human reading pattern of a document and outputs a unique semantic representation for every text span 
irrespective of the content type they are found in. 
We base our architecture on a graph representation of the structured text,
and we demonstrate that not only can we retrieve semantically similar information across documents but also that the embedding space we generate captures useful semantic information, 
 similar to language models that work only on text sequences. 
}

\documentclass[11pt, a4paper, twocolumn]{article}
\usepackage{xurl}
\usepackage[square,numbers]{natbib}
\usepackage{abstract}

\usepackage{lipsum}
\usepackage{amsmath}
\usepackage{amssymb}
\usepackage{graphicx}
\usepackage{cuted}

\usepackage{hyperref}
\hypersetup{colorlinks=true, urlcolor=blue, linkcolor=blue, citecolor=blue}

\begin{document}


\twocolumn[
  \begin{@twocolumnfalse}
    \maketitle
    \begin{abstract}
      \abstractText
      \newline
      \newline
    \end{abstract}
  \end{@twocolumnfalse}
]


\section{Introduction}

When reading a document, people naturally switch between very different reading patterns. 
The pattern for reading a table generally involves a sort of reverse L-shaped scan over to the row and up 
to the column label whilst for a paragraph the order is top-to-bottom, left-to-right in written English. 
A person can also map information collected by different reading patterns into the same semantic space. 
For example, one can relate a number representing gross-earnings-per-share read from a table to the same figure written in a paragraph.\\

In principle, this is just a generalised statement of the distributional hypothesis that underpins modern NLP approaches: 
 words that occur in similar contexts purport similar meanings \cite{Harris1954}. 
In the NLP literature the "context" has typically been understood as a sequence of word tokens. 
In a realistic model of reading, we believe the context must be variable and a person chooses to apply different context patterns based on page semantics.\\

There has been great progress in the Natural Language Processing (NLP) literature by treating different reading patterns as separate problems. 
For paragraphs, language models like BERT \cite{Delvin2018} can be used to generate vector representations of text as inputs into downstream NLP tasks. 
For more complex reading patterns like tables, methods like \cite{Paliwal2019} leverage computer vision models to identify tables and then other algorithms 
to further decompose the tables into a useable data-structure.\\

To build systems that can read documents like a person, however, we need to go further than this. 
On addressing a page, such systems need to work out the required reading pattern for a specific content type (paragraph, table, etc…) 
and then create a representation that is interoperable with representations generated with different reading patterns on other content types.\\

There are large benefits to pursuing this type of holistic document reading model. 
Most business documents, research papers, slide decks etc., require readers to deploy multiple different reading styles to understand the nuance of the content. 
Needing to build 2 or more systems to read different parts of the document is expensive and means numeric content in tables is then kept separate from narrative content. 
One can write a 3$^{rd}$ piece of software to merge the datasets together again ex-post-facto, but this is clearly sub-optimal and not what a person does.\\

The most relevant research in this area has focused on building systems that can complete named entity detection (NED) tasks on documents 
like shopping receipts \cite{SROIE} or scanned company filings \cite{Kleister2021}. 
These images will often be poor quality and the text may be misaligned so optical character recognition can fail and/or the text can come out in a jumbled order. 
As a result, researchers need to include extra visual cues in their models to build in some redundancy 
and whilst none of these methods addresses reading strategy directly, they provide interesting insights as to how we might proceed.\\

LayoutLMv2 \cite{Xu2020} and LAMBERT \cite{Garncarek2020} both modify the positional encoding that BERT uses for text sequences to include information about the 2D positioning 
of text tokens on a page. The positional encoding then works the same way as BERT so the model can increase or decrease emphasis on tokens 
based on either their relative positions to a central token, or their absolute position on the page. 
StructText \cite{Li2021} enriches the input by grouping tokens into contiguous sequences then having both token 
and sequence level representations as inputs to a BERT-like model. 
This aims to capture the fact that there may be multiple levels of semantic grouping on a page, token, line, paragraph etc.\\

The work of Y. Hua $\&$ al.\cite{Hua2020} takes a slightly different approach: their model uses unmodified BERT vectors 
but then builds a graph of the page where text tokens have a directed edge to other tokens in the same line and the line immediately above them. 
This graph is used as the input into a graph attention network (GAT,\cite{Velickovic2018}) which can be trained on specific named entity recognition tasks.\\  

These methods perform well on benchmark NED tasks so we can conclude the inclusion of extra context in this way is valuable, 
however, the idea of reading strategy is somewhat more refined than this. We would expect, for example, that given a number in a table, 
a good holistic reading model would be able to highlight the row and column headers and perhaps the table title and other information needed to identify the semantic meaning of the number. 
This information can exist at a variable distance from the number and associative visual information. 
For example, similar alignments of numbers leading up to a text span in a columns play key roles in a person being able to identify the correct reading pattern. 
This more complex idea of reading strategy cannot be modelled in existing architectures.\\

In this paper we introduce a model that given a piece of text in a document with paragraphs, lists, tables etc., 
can identify the correct reading pattern to apply, and then uses it to build an embedding vector which usefully represents the semantic meaning of the text. 
Following a similar approach to \cite{Mikolov2013}, we evaluate this model on a set of publicly available financial reports, 
using several tests to show that our vectors draw information from sensible reading patterns given different page contexts (tables, lists, paragraphs etc.) 
and that vectors that represent text with similar semantic meanings cluster together in the embedding (Euclidian) space.   

\section{Methodology}
The basic approach we take here is to model a page in a document as a directed graph, 
where each vertex in the graph is a sensible unit of text and each edge is a potential way a person's eye may move 
from one unit of text to another when reading the document. 
This graph is meant to model all the potential reading patterns a person is likely to deploy. 
We then use a Graph Neural Network with self-attention mechanism (through transformer encoder layers) 
to try and learn the correct reading pattern from the set of potential reading patterns when creating the vertex representations. 
If our model works, the attention weights should activate/deactivate vertices that sit on sensible reading pathways. 
The output of the model is an embedding vector for each vertex that is a weighted average over the context the model determines 
as the correct reading pattern for that situation.  

\subsection{Graph Definition}
\label{graph_def_section}
When defining our graph, the first task is to generate “sensible units of text” we call “spans”. 
To generate spans, we first arrange text tokens into lines based on the bottom coordinate of their bounding box. 
We then cut these lines where we think the whitespace separation between tokens indicates they belong to separate columns. 
We use an optimisation algorithm to do this but note that other authors do something similar with CNN architectures \cite{Hua2020}.\\ 

Once we have defined spans, we draw up to 4 edges between each span based on the 4 possible movements 
your eye can make to when addressing the page at that span: up, down, right, and left. 
We define each edge as the connection to the closest neighbour span when moving in that direction on the page. 
The result is a graph $\mathcal{G}$, with vertices $\mathcal{V}$, where each $v_{i}$ has a neighbourhood $e(v_{i})$ with up to 5 vertices in it, 
the vertex above, below, left, right and the vertex $v_{i}$ itself.

In most cases, we find that if $v_{i}$ has a directed edge to $v_{j}$ then typically $v_{j}$ 
will have the reverse edge back to $v_{i}$ and so the graph is ‘mostly’ undirected. 
However, there are important cases where this is not true, for example, if a title in centrally aligned on the page 
with two columns of text underneath it, reading order dictates you read the column on the left first in written English, not both simultaneously.\\ 

Although we focus on PDF files in this article this approach allows us to represent information for any kind of document 
regardless of their layouts (A4 landscape, portrait etc.,) or their type (PDF, .xlsx, .docx etc.,) 
and so generalises well to other document understanding use cases. 

\subsubsection{Vertex feature vector generator}
The next step in graph creation is to convert the text spans into feature vectors that serve as the inputs into our neural network. 
There are many language modelling methods available in the literature for converting text tokens into fixed length vector 
e.g., BERT, Word2Vec, continuous bag of words (CBOW) etc. In our use case there are a couple of added complications in deploying these models:\\
First, our documents contain many tables as well as text and, in the tables, there are many numbers and dates that occur in spans on their own. 
Language models are well known to represent numbers poorly \cite{Thawani2021}.\\
Secondly, they also contain financially specific language that studies such as \cite{Loughran2011} have shown differs significantly from standard English. 
For example, words such as ‘liability’ have a completely different meaning in financial documents to the common English usage.\\

We initially experimented with using BERT$\_$base and a CBOW model that we trained on a corpus of financial documents. 
To create a span vector, we tried averaging over all the vectors either model produced for a span as well as using the [CLS] token in BERT$\_$base 
following the work of H. Choi $\&$ al. \cite{Choi2021}. We also tried different strategies for masking numbers and dates.\\

In the end we found that we got the best performance by masking numbers by their magnitude (i.e., tens, hundreds, thousands, and millions) 
and by adding extra keywords to indicate whether the number is a currency, a percentage or simply a basic quantity. 
For dates we masked them to tokens for day, month, year, and quarter. We chose to use BERT instead of CBOW for this study because it can take into account out of vocabulary tokens 
and therefore is more flexible. 
We use BERT$\_$base model for this set of experiments but we have no doubt that results could be boosted by using a financial fine-tuned model.

\subsubsection{Relative position information}

In addition to vertices intrinsic features, we compute the relative position of vertices on the page using Manhattan distance \cite{ManhattanD}. 
The geometric nature of our graph means that edges are drawn between vertices along two perpendicular axes 
allowing us to calculate the relative Manhattan distance by just counting the hops between vertices. 
We store this information in the matrices $P_{vert}$ and $P_{hor}$ for the vertical and horizontal axes respectively 
with $P_{vert}$ and $P_{hor} \in \mathbb{Z}^{N\times N}$, N being the number of vertices in the graph. 

\subsection{Graph Neural Network}

When choosing a neural network architecture to run on a graph \cite{Chami2020}, we considered that Graph Convolutional Networks (GCN) are not suitable for the problem at hand because the reading pattern 
is not uniformly distributed in the vertex neighbourhood. For example, column headers may be many hops away from the vertex of interest 
in a single direction. Attention-based graph networks are more suited to this problem because 
they allow messages to be passed on longer distances by altering the attention weights in the network. 
We opted for a transformer \cite{Vaswani2017,Delvin2018,Dwivedi2020} as the attention mechanism for two main reasons:\\

\indent$\bullet$ We experimented initially with the flat attention mechanism form \cite{Velickovic2018} 
and found that as the attention is applied elementwise it couldn’t learn to adapt 
to sequences of hops like ‘number-number-column header’ rather it just filtered out particular elements of the input vector. 
The transformer mechanism assigns attention weights as a function of all elements in the input vector and its neighbours which worked better.\\

\indent$\bullet$ We felt that a pure message passing mechanism might be too brutal a filter when considering a single span. 
For example, given a number in a table with a reverse L-shaped reading pattern, 
this would mean that the model can only ‘perceive’ vertices directly on this traversal. 
It seems likely a person reading a table also considers some information from other vertices in their peripheral vision 
or remembers vertices like the title of the page and merges this information in her internal representation of the number. 
We want an architecture that is flexible enough to allow us to experiment with the interplay between pure message passing 
and other types of information flow via additional edges and positional encoding.  

\subsubsection{Model architecture}

\begin{figure*}[h]
  \begin{center}
    \includegraphics[width=5in]{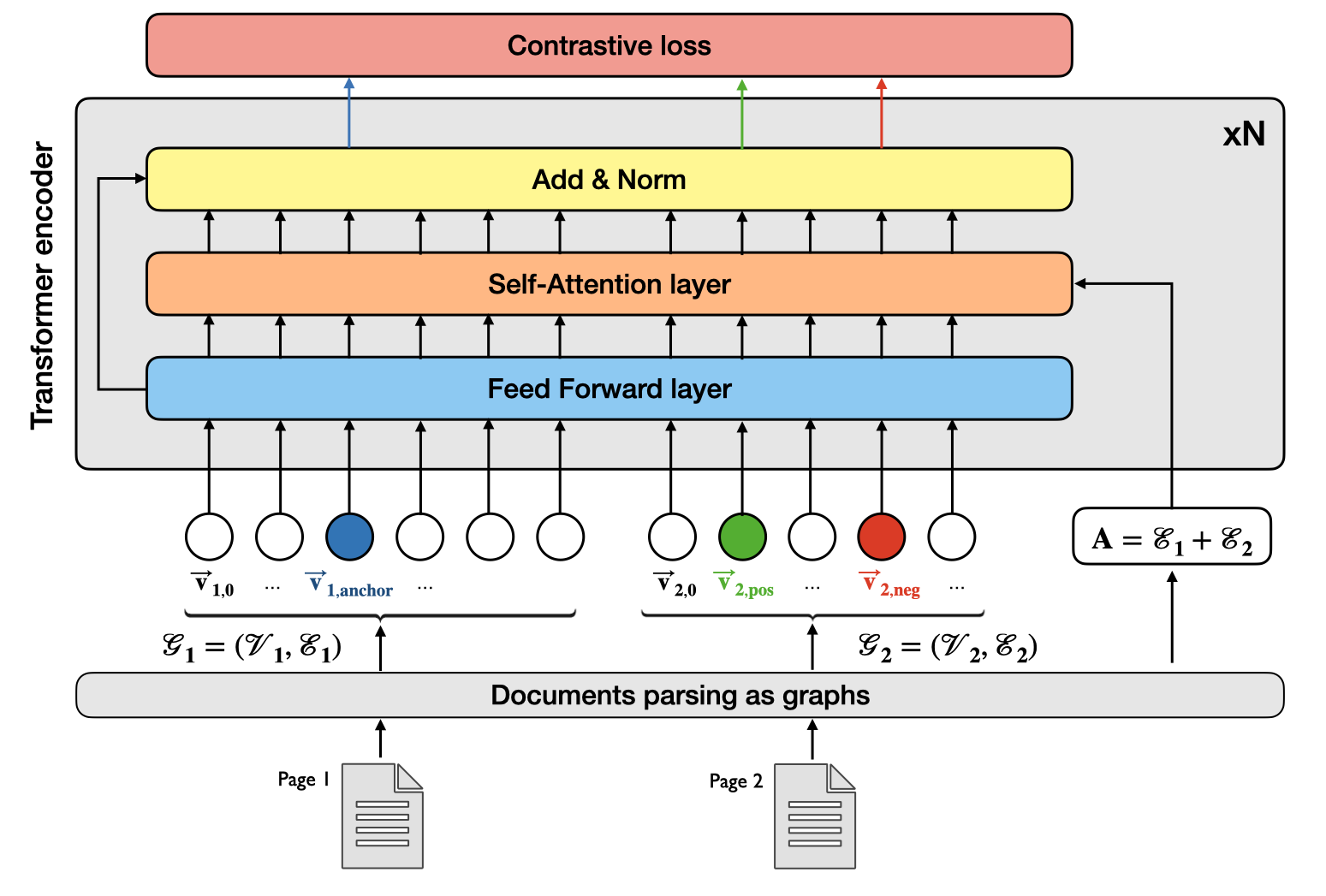}
  \end{center}
  \caption{Architecture of the Graph Neural Network for training. Each vertex goes through 
  a series of N$_{layers}$ encoding layers that contains self-attention.}
  \label{fig_arch}
\end{figure*}

Our model input is a graph $\mathcal{G}$, with a set of N vertices $\mathcal{V}$  
with each $v_{i}$ having a neighbourhood $e(v_{i})$ which is a set of up to 5 vectors in $\mathcal{V}$ 
as defined in section \ref{graph_def_section} above. An adjacency matrix $A$ encodes the edges between the vertices. 
The first layer of the GNN takes $\mathcal{V}$ and produces a set of modified vectors $\mathcal{V}'$ with the same dimension as $\mathcal{V}$ 
by passing each $v_{i}$ as the query and $e(v_{i})$ as the key into a transformer encoder (figure \ref{fig_arch}) and outputting $v_{i}'$ as an attention weighted average over the neighbourhood $e(v_{i})$. 
So, for each vertex, for each forward pass, we calculate:

\begin{equation}
  \vec{v}_{i} = \sum_{j}^{N}\alpha_{ij}V\vec{v}_{j}
\end{equation}
with 
\begin{equation}
  \alpha_{ij} = \frac{1}{\sqrt{d}}\frac{A_{ij}(Q\vec{v}_{i} \cdot K\vec{v}_{j})}{\sum_{j}^{N}A_{ij}(Q\vec{v}_{i} \cdot K\vec{v}_{j})}
\end{equation}

where $\alpha_{ij}$ is the attention coefficient of $v_{i}$ to $v_{j}$, $Q$,$K$ and $V$ are query, key and value matrices respectively, 
$A_{ij} \in [0,1]$ is the adjaceny matrix element for the i,j pair. $d$ is the features dimension of the vertices vectors. For subsequent layers we then take $v_{i}'$ as input with $e(v_{i}')$ now being the 
neighbourhood of vectors in $\mathcal{V}'$ and so on. We decided to evaluate the same GNN architecture in two different configurations:\\

\indent$\bullet$ A message passing network case where edges are limited to the 5 nearest neighbours: ($A_{ij} = 1$ if $|P_{ij,vert}|\leq1$ or $|P_{ij,hor}|\leq1$) 
as this allow us to preserve the geometry of the document graph as an inductive bias and we want to model how information flows from one span of text to another over the graph. 
In this architecture information can flow from vertices that are very far away from each other with the maximum accessible distance 
a function of the number of layers in the model but at the cost of an exponential damping of information flow.\\  
\indent$\bullet$ A regularised model to compensate for long distance information damping in the purely message passing architecture. 
This version  includes an arbitrary number of edges ($A_{ij} = 1$ if $|P_{ij,vert}|\leq x$ or $|P_{ij,hor}|\leq x$ with $x\in\{5,8\})$. 
Because the geometry of the graph is partially lost in this configuration, we apply the following regularisation inside the transformer expression:  

\begin{equation}
  \vec{v}_{j} =
    \begin{cases}
      \vec{v}_{j} & \text{if $\sqrt{P_{ij,vert}^{2}+P_{ij,hor}^{2}}\leq 1$}\\
      \vec{0}  & \text{otherwise}
    \end{cases}       
\end{equation}

It dictates a uniform distribution of attention on a neighbourhood of order $>$ 1 
that is anti-correlated with the attention coefficients from the vertex nearest neighbours. 
Applying positional encoding and or distance regularisation can only be done in a relative way as graphs don’t have a fixed shape 
(unlike images) and padding the graph with empty vertices (such as in BERT sequences) is not an option for computation time and memory reasons. 
Introducing this information inside the self-attention layers to the key vector is therefore the natural solution. 
Discussion about relative positional encoding in transformers can be found in this work \cite{Chen2021}. 

\subsubsection{Training objective}

Typically, language models like CBOW and BERT are pre-trained on an unsupervised task where tokens in a sequence are masked, 
and the training objective is to predict the masked tokens. This does not work in our case as for two reasons: 
1) to mask tokens effectively we need to know what tokens are important in advance of calculating the loss. 
In our case, reading pattern is learned dynamically so we do not have this information. 
2) Many of our tokens are numbers that have little semantic information and so predicting them is not challenging.\\

To get around this issue we trained the model in a supervised manner using a contrastive loss function \cite{Zhu2020}. 
This allows us to identify information we think should be closer together/further away in embedding space and force the model 
to learn the reading patterns that make this happen. 

We train our model using pairs of graphs $\mathcal{G}_{1}=(\mathcal{V}_{1},\mathcal{E}_{1})$ and $\mathcal{G}_{2}=(\mathcal{V}_{2},\mathcal{E}_{2})$ that 
we bind in single input graph $\mathcal{G}=(\mathcal{V}_{1}+\mathcal{V}_{2},\mathcal{E}_{1}+\mathcal{E}_{2})$ (figure \ref{fig_arch}). 
We do not need to add a separation vertex between the graphs as no edges exists between vertices of graphs 1 and 2, 
therefore no information can flow between the two graphs. We label pairs of semantically similar vertices ($v_{anchor}\in\mathcal{V}_{1}$ and $v_{pos}\in\mathcal{V}_{2}$) 
between graphs and vertex $v \in\mathcal{V}_{2} / v_{pos}$ most similar to $v_{anchor}$ is chosen as $v_{neg}$ after each training iteration. 
We use the following definition of contrastive loss:

\begin{multline}  
  \mathcal{L} = ||\vec{v}_{anchor}-\vec{v}_{pos}|| \\+ \max(0,m-||\vec{v}_{anchor}-\vec{v}_{neg}||)
\end{multline}

with $m$ the margin hyperparameter. By applying a max() function on the second part of the formula 
we intend to decorrelate the optimisation of the positive distance from the negative one. 

\subsubsection{Training and validation datasets}

We use public financial reports from various companies to train and evaluate our model. 
For each company we downloaded two quarterly reports (Q2 and Q3 2020) from the investor relations section of publicly listed companies’ websites. 
These reports are interesting to us because, whilst there is a set of information that they legally must contain regarding company performance, 
there is a large variety of other information and different formatting types (paragraphs, lists, tables etc.,). 
As a result, documents have some shared information but presented in different reading contexts.

The training and validation corpus consists of 70 pairs of pages graphs, each of them containing an average of 100 vertices and representing a training batch. 
We filtered out batches containing only a few vertices but still have variable batch sizes for the training. 
The total number of labelled pairs of vertices is $\sim$7000. The Model hyperparameters are N$_{layers}$=8 feed-forward + transformer layers of size 360 each 
(equivalent to the feature size, no compression is made) and it was trained for 400 epochs. 
We also used 5-fold cross-validation to obtain the final validation loss and accuracy. 

The model was subsequently tested on a corpus of 10 pairs of pages and the results are presented in the next section. 

\section{Experiments}

We have conducted a set of experiments to demonstrate that our model has the ability 
to capture 2D reading order by showing 2D reading patterns in tables that it creates a the semantic meaningful embedding space.

\subsection{Similar vertex pairing task}

Our first task is a straightforward out-of-sample test of whether our model can find vertices we have labelled as pairs in our corpus. 
To do this we take every labelled vertex and calculate the Euclidean distance between its vertex embedding and every other vertex 
in the same document its pair can be found in. So, the test is out of all the vertices in document 2, 
can the model match the one in document 1 with its most similar counterpart. 

Specifically, we take 2 documents and calculate their vertices embedding sets $\mathcal{V}'_{1}$ and $\mathcal{V}'_{2}$ for document 1 and 2 respectively 
using the same trained GNN model and where document 2 has K vertices. Then assuming we have a labelled vertex $v’_{1,i}$ with its pair $v’_{2,i}$, 
we first calculate a vector $d_{i}\in \mathbb{R}^{K}$ of distances between $v’_{1,i}$ and every other vertex in $\mathcal{V}'_{2}$ using 

\begin{equation}
  d_{i,k} = || \vec{v} ’_{1,i} - \vec{v} ’_{2,k}||\hspace{0.5cm}\forall\hspace{0.125cm} k\in K
\end{equation}

For corpus C of out of sample documents with M labelled vertices in total we define the total score as 

\begin{equation}
  score(C) = \frac{1}{M} \sum_{i}^{M} 
    \begin{cases}
      1 & \text{if $\arg\min(d_{i})=i$}\\
      0 & \text{if $\arg\min(d_{i})\neq i$}
    \end{cases}       
\end{equation}

We tested the on a range of different span types i.e., from paragraphs, lists, and tables. 
For general text in paragraphs the model performs extremely well but we note that other language modelling approaches 
based purely on text sequences would be expected to perform as well on this type of task, 
so we focus our analysis on the performance of our model on tables. 


We also have evaluated one of the state-of-the-art language models with 2D structural information encoding, 
LayoutLMv2\cite{Xu2020}, for comparison. To run this experiment on LayoutLMv2 we used its pretrained version available on Huggingface. 
Since it outputs an embedding vector for each word token on a page, we reconstructed span embeddings 
by averaging the tokens embeddings included in the span using our existing graph model to determine what tokens should be in what span.  

\begin{figure}[h]
  \begin{center}
    \includegraphics[width=3in]{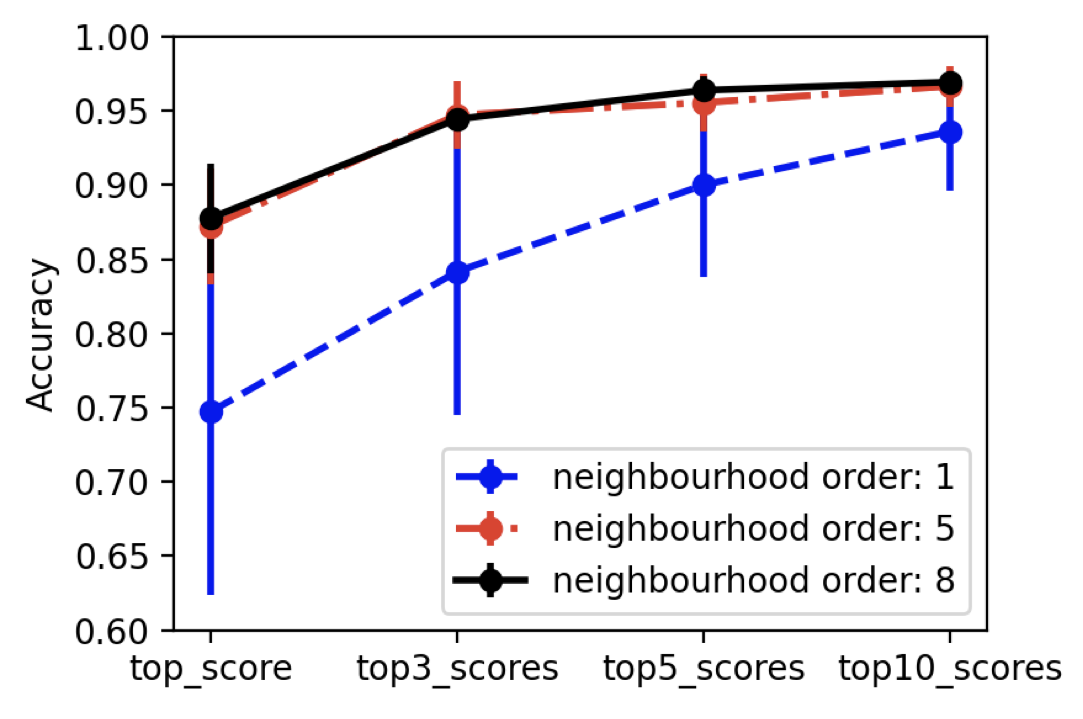}
  \end{center}
  \caption{Comparison of model accuracies for finding matching vertex in top 1, 3, 5 and 10 predictions. 
  1$^{st}$ order neighbourhood corresponds to results for the unregularized message passing model.}
  \label{fig_acc}
\end{figure}

We started our study by comparing accuracies of the pure message passing model versus the regularised one (figure \ref{fig_acc}). 
Both 5 order and 8 order neighbourhood regularised model exhibit significant improvement in the similarity matching task 
confirming that pure message passing results in information damping and reduces the efficiency of the model. 
Higher order neighbourhoods seem to marginally increase the probability of finding the right match whether 
considering the top 1, 3, 5  or 10 predictions.\\

\begin{table*}
  \begin{center}
  \begin{tabular}{|c|c|c|c|c|}
    \hline
    $\#$ of columns & LayoutLMv2 & Our model & Our model  & Our model\\
     & & 1$^{st}$ order & 5$^{th}$ order & 8$^{th}$ order \\
    \hline
    2 & 0.58$^{*}$ & 0.747 & 0.872 & 0.878\\
    \hline
    3 & - & 0.729 & 0.879 & 0.884\\
    \hline
    4 & - & 0.702 & 0.882 & \textbf{0.89}\\
    \hline
  \end{tabular}
  \caption{Vertex similarity retrieval accuracies for different table sizes. $^{*}$LayoutLMv2 has only be tested for tables with 
  $\leq$2 columns because of the limitation of the model input tokens sequence (512).}
  \label{table_accuracies}
  \end{center}
\end{table*}

Table \ref{table_accuracies} shows how results exhibit high accuracy in the regularised models. 
The accuracy of the unregularized model also slightly reduces as the table size increases. 
We interpret this drop as a limitation of pure message passing GNN because if there is any noise 
in the statistical relationships between vertices in the graph then this will be multiplied by the number of hops. 
This highlights the exponential damping effect where the deeper the vertex lies in the table the harder 
it is to distinguish it from surrounding table vertices. 

LayoutLMv2 model performance for this task is significantly below ours. Which we think can be explained by 3 main factors:\\
1) LayoutLMv2 has no concept of a span as it embeds only word tokens. We think this is a key issue as the vertical typesetting 
of a page into columns is based on the alignment of spans and not individual word tokens. Without having a concept of a span, 
models can’t capture this important property of page semantics.  

2) Our model is specifically trained to optimise unique representations of spans and therefore it is forced 
to find information on the page that can generate a vector for a number in a table that is unique from other numbers, 
even if the tokens themselves are the same. In contrast methods like LayoutLMv2 focus on predicting individual 
word tokens which in the case of numbers in tables is a somewhat arbitrary task (as discussed above).  

3) We masked every numerical span before generating embeddings whereas LayoutLMv2 and other BERT based 
sequential models tokenize specific numbers before generating the embeddings. 
As mentioned in a previous section these numeric representations are known to be problematic \cite{Thawani2021}. 

In the following experiments we use the regularised model with a neighbourhood of order 8 as our reference model.

\subsection{Analysis of semantic usefulness of embeddings}

\begin{figure*}[h]
  \begin{center}
    \includegraphics[width=6in]{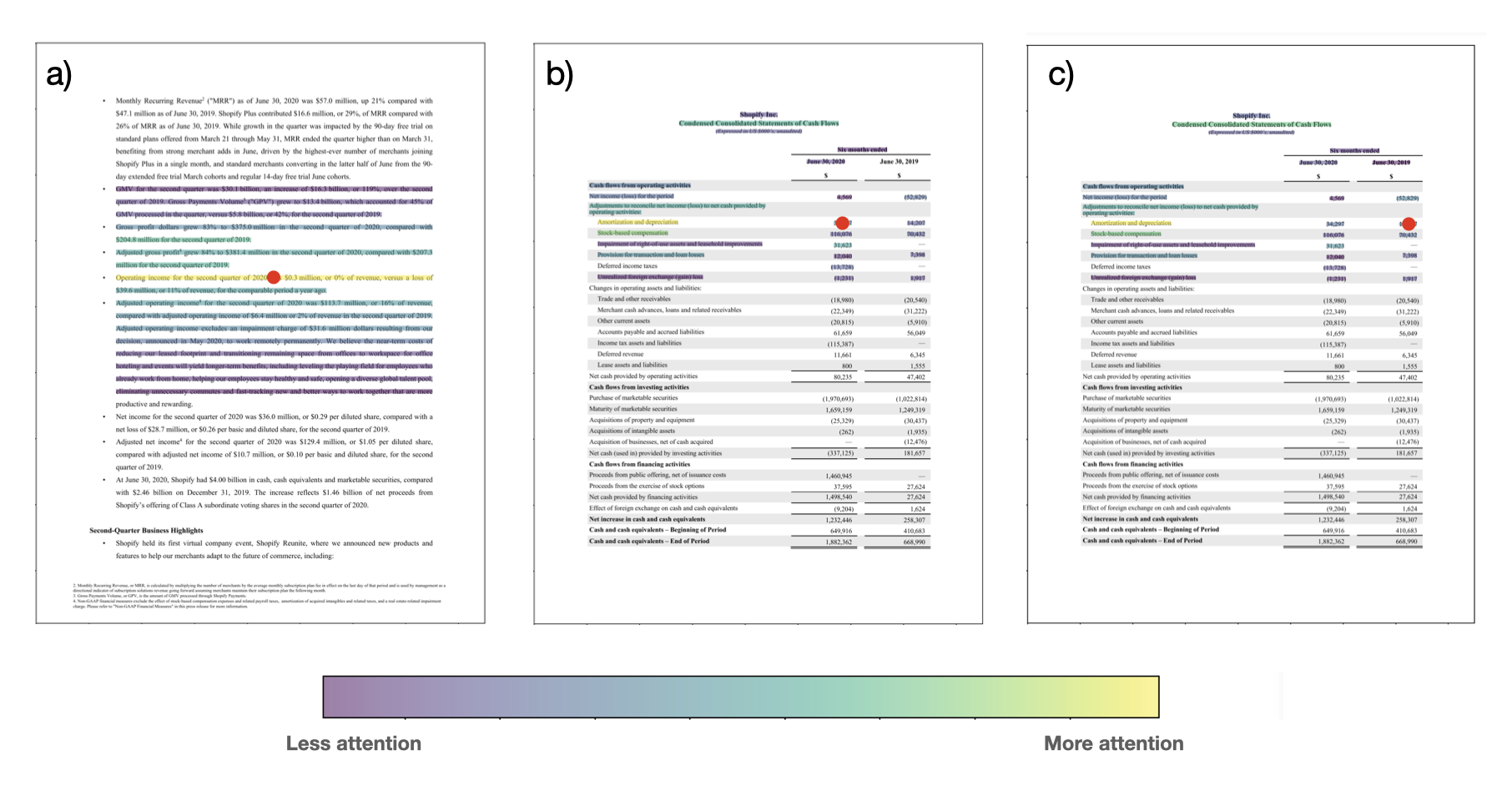}
  \end{center}
  \caption{ Projected attention pattern on documents for 3 examples datapoints. a) for a page with only paragraphs. 
  b) and c) for a table document. Spans of text not highlighted in the figure contribute only a fraction of 
  a percent to the attention and therefore are not considered here.}
  \label{fig_attn}
\end{figure*}

To check if the model really is learning the expected reading pattern, rather than fitting to some arbitrary features, 
we generated attention overlays onto pages of our documents using the attention rollout method proposed by S. Abnar $\&$ al. \cite{Abnar2020}. 
Figure \ref{fig_attn} shows the attention rollout maps overlayed on2 different pages for 3 different vertices. 
In each case the red dot is the vertex of interest and the colour maps on the page show the different attention weights 
for each vertex the model has used to generate the vertex of interest’s vector. 

We clearly see from these maps the model is learning the expected reading pattern. 
Panel a) shows how when a span in a paragraph is selected the model has an exponentially decaying attention map moving above and below the span in the paragraph. 
Panel b) and c) show examples of table cells and how the model picks up row and column header information, 
concentrating most on the row label as this tends to contain the most semantically important information 
for identifying the semantic meaning of the number. Examples of attention maps for a single page with different regularisation orders 
can be found in the appendix.\\

\begin{figure*}[h]
  \begin{center}
    \includegraphics[width=6in]{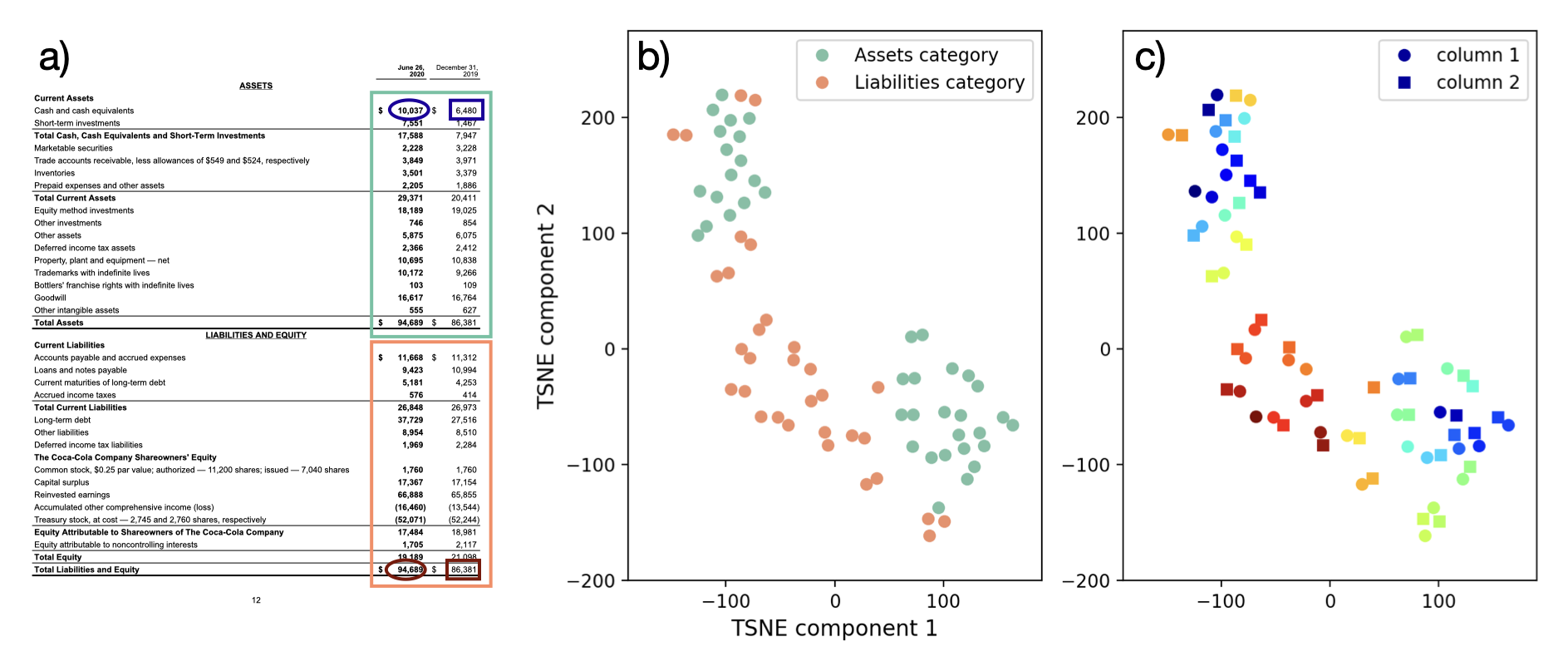}
  \end{center}
  \caption{ T-SNE Projection of vertices embedding vectors in 2D. a) Document page example, 
  the different row label sections that the datapoints are part of are highlighted in green and orange. 
  For each row we looked for the embeddings in both columns b) Projected distribution of table values per table section. 
  c) Same projection with different color for each row and shape of the marker to distinguish the two columns}
  \label{fig_cluster}
\end{figure*}

We also researched how different embeddings clustered together. We would expect, for example, vertices 
from similar columns or rows in tables to cluster together in embedding space. 
To analyse this, we projected the embeddings in 2D using T-SNE \cite{Vandermaaten2008} and plotted vertices clusters.  
Figure \ref{fig_cluster} shows table vertices from same section clustering together meaning that vertices embeddings contain information about their row.  
Moreover, within sections one can see that vertices cluster by pair corresponding to the pair of columns in the table. 
This is a clear indication that the model reads column and row information to create a unique semantic representation of each vertex. 
The vertices appear to be initially clustered by row section and then row clusters can be cut into column components suggesting 
that the model prioritise row information for the vertex embeddings in this example.\\

Finally, we researched compositionality. In the original Word2Vec paper \cite{Mikolov2013} one of the key findings was that if $\vec{w}_{king}$ represents 
an embedding for the word “king” generated by the Word2Vec model then $\vec{w}_{man}-\vec{w}_{king}+\vec{w}_{woman}$ would generate a vector closest in embedding space to $\vec{w}_{queen}$. 
The intuition behind this experiment is that presumably the words “man” and “woman” will occur in similar sequences in general, 
however, there is likely to be some language that is idiosyncratically gendered and so “man” will be pushed towards other male words and “woman” towards other female words. 
Similarly, “king” and “queen” will occur in similar sequences but also both be affected by the same idiosyncratically gendered language. 

The result is that $\vec{w}_{man}-\vec{w}_{king}$ effectively suppresses idiosyncratic maleness as both vectors will be in similar positions 
relative to more male words. Similarly, $\vec{w}_{woman}-\vec{w}_{queen}$ cancels out idiosyncratic female word correlations and so the resulting vectors 
are similar once gendered language is removed.  

Column clustering in figure \ref{fig_cluster}.c hints at the possibility of doing an analogous task. 
In the table example below (table \ref{compo_table}) we would expect that there is idiosyncratic information about “earnings” or “costs” contained in the table and surrounding text. 
So similarly, (Earnings 2019)-(Earnings 2020)+(Costs 2019) should generate a vector close to the embedding for (Costs 2020). 

\begin{table}
  \begin{tabular}{|c|c|c|}
    \hline
     & 2019  & 2020\\
     \hline
     Earnings & (earnings 2019) & (earnings 2020)\\
     \hline
     Costs & (costs 2019) & (costs 2020)\\
    \hline
  \end{tabular}
  \caption{Example of table for compositionality experiment}
  \label{compo_table}
\end{table}

To test this, we evaluated the truthfulness of the following equation:
\begin{equation}
  \vec{v}_{l,i} - \vec{v}_{k,i} + \vec{v}_{k,j} = \vec{v}_{l,j}
  \label{eq_compo}
\end{equation}

where $\vec{v}_{k,i}$ is the value in the k$^{th}$ column and the i$^{th}$ row and $\vec{v}_{l,i}$ is in the l$^{th}$ column and the i$^{th}$ row with k$<$l.
We applied equation \ref{eq_compo} to 3 tables of our test set with variable row lengths (20, 32 and 35 rows). 
This generated 87 ($\vec{v}_{l,i} - \vec{v}_{k,i}$) vectors which we then applied to every row from their respective tables corresponding to 
2649 applications of the equation to find exact matches like in the earnings example above. 
We recorded an exact match success rate of 78.8$\%$, demonstrating that the embedding space exhibits compositional properties. 

\section{Conclusion}

Creating useful distributed representations of word sequences has been a key driver of success in benchmark NLP tasks. 
However, as we have discussed in this article, sequences are only one of the patterns a person deploys when reading most documents. 
If we are to replicate the success of models like Word2Vec, Glove or BERT across the full range of reading scenarios we need to build models 
that generate distributed representations for all these patterns.  

In this article we presented a method that can approximate realistic human reading patterns across all the typical document reading scenarios. 
Our method works by transforming structured text into a graph of spans, then training a Graph Neural network to represent 
semantically similar spans with similar vectors using a contrastive loss objective. 

We benchmarked our model on table content embeddings because they constitute the most challenging part of the problem 
for standard sequence-based language models. We demonstrated that our learnt reading pattern is a good first approximation of the human one, 
allowing us to retrieve similar datapoints in tables from different documents with an accuracy of 89$\%$. 
We also showed that, in analogy to language models, the embedding space we create exhibits useful semantic properties that allow 
similar text clustering and compositionality. 

Our contribution extends the literature on language models to the full text of documents, particularly business or academic documents, 
that contain tables and lists intermingled with paragraphs and narrative text. 
This paves the way for exciting NLP tasks such as question answering, Named Entity Recognition and sentiment analysis 
to be explored fully in such documents. 

\section{Acknowledgments}

We would like to thank the analysts in the team who contributed setting up the dataset and analysing the model results: 
Victoria Li, Hannah Lias, Sachie Tran, Xinyang Jin, Seoin Chai, Monica Macovei, Timothy Cardona and Serges Saidi.


\bibliographystyle{unsrtnat}
\bibliography{paper}

\newpage
\onecolumn
\section*{Appendix}

\begin{figure*}[h]
  \begin{center}
    \includegraphics[width=5in]{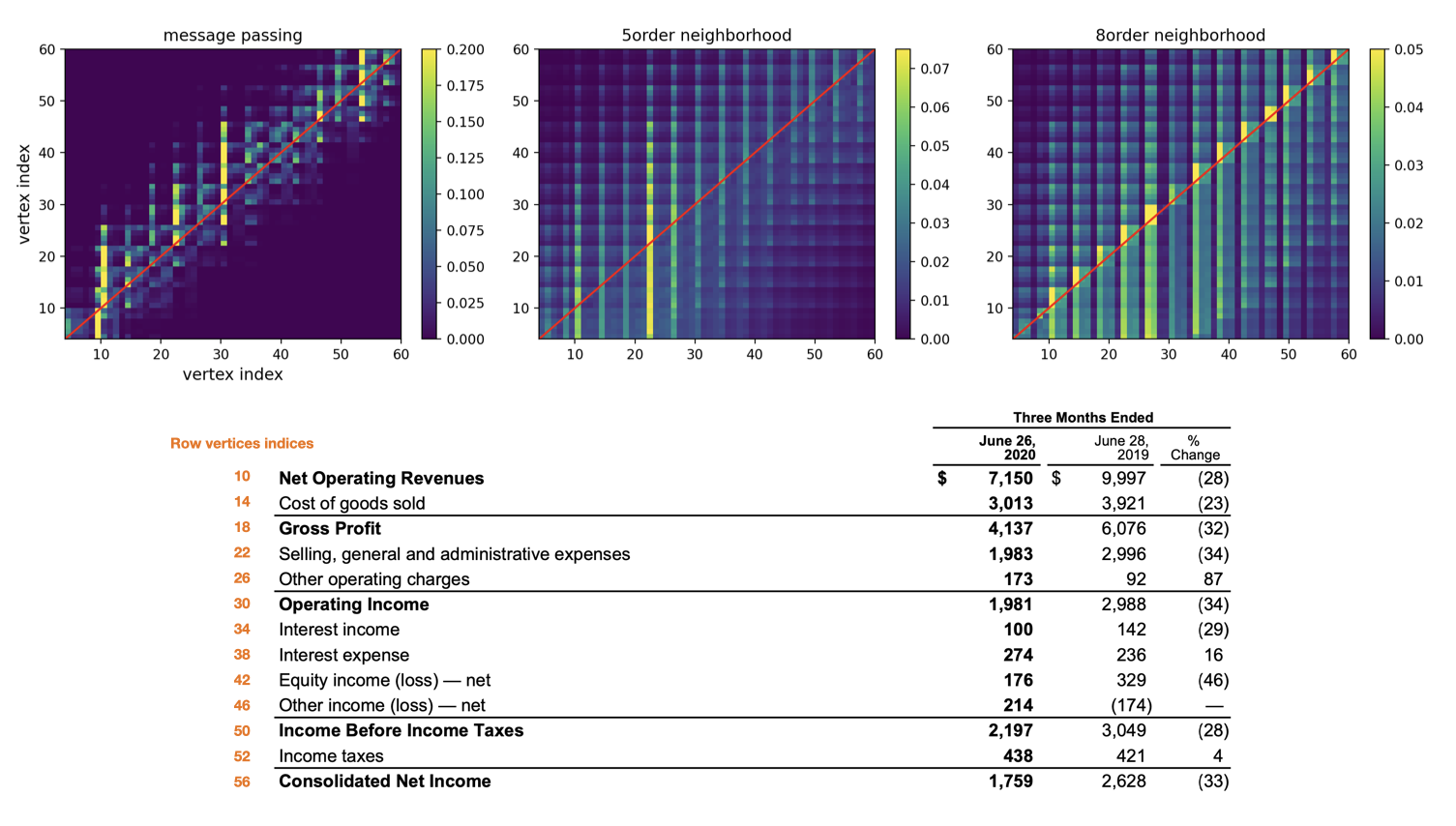}
  \end{center}
  \caption{A single page table attention rollout maps for the three tested model configurations.}
  \label{fig_appendix}
\end{figure*}
Figure \ref{fig_appendix} shows the difference of attention on the row labels for the models we tested. 
We clearly observe the exponential damping of attention in the message passing case. 
Moreover row label attention appears peculiar. Some labels contribute strongly for values on different rows and some 
others have competing attention with numbers in the table (for vertices of indices 35 to 45). 
It is in accordance with the worst vertex similarity matching task observed for this model. 
The configuration including 5 orders of neighbourhood with regularisation exhibit a more structured pattern of 
attention with most of it on the row labels nevertheless those labels seem to contribute to  multiple row values representation. 
Finally the 8 order neighbourhood model seems to capture maximum attention from the labels on the same row as 
their corresponding values and a general context information given by the uniform attention to all other nodes. 
We note as well that for all three models the attention to the row labels present some asymmetry toward upper 
left triangle translating into more relative attention to vertices on the left and above which is 
what is expected for a reading pattern in a table.

\end{document}